\documentclass[runningheads]{llncs}
\usepackage{graphicx}

\usepackage{tikz}
\usepackage{comment}
\usepackage{amsmath,amssymb} 
\usepackage{color}
\usepackage{booktabs}
\usepackage{hyperref}

\DeclareMathAlphabet{\mathpzc}{T1}{pzc}{m}{it}

\begin{document}
\pagestyle{headings}
\mainmatter
\def\ECCVSubNumber{6}  

\title{Graph Neural Networks for 3D Multi-Object Tracking\vspace{-0.3cm}} 

\titlerunning{ECCV-20 submission ID \ECCVSubNumber} 
\authorrunning{ECCV-20 submission ID \ECCVSubNumber} 
\author{Anonymous ECCV submission}
\institute{Paper ID \ECCVSubNumber}

\titlerunning{Graph Neural Networks for 3D Multi-Object Tracking}
%

\author{
Xinshuo Weng \and
Yongxin Wang \and
Yunze Man \and
Kris Kitani}
\authorrunning{X. Weng et al.}
%
\institute{Robotics Institute, Carnegie Mellon University \\
\email{\{xinshuow, yongxinw, yman, kkitani\}@cs.cmu.edu}}
\maketitle

\begin{abstract}
\vspace{-0.65cm}
3D Multi-object tracking (MOT) is crucial to autonomous systems. Recent work often uses a tracking-by-detection pipeline, where the feature of each object is extracted independently to compute an affinity matrix. Then, the affinity matrix is passed to the Hungarian algorithm for data association. A key process of this pipeline is to learn discriminative features for different objects in order to reduce confusion during data association. To that end, we propose two innovative techniques: (1) instead of obtaining the features for each object independently, we propose a novel feature interaction mechanism by introducing Graph Neural Networks; (2) instead of obtaining the features from either 2D or 3D space as in prior work, we propose a novel joint feature extractor to learn appearance and motion features from 2D and 3D space. Through experiments on the KITTI dataset, our proposed method achieves state-of-the-art 3D MOT performance. Our project website is at \url{http://www.xinshuoweng.com/projects/GNN3DMOT}.
\vspace{-0.15cm}
\keywords{multi-object tracking, graph neural networks}
\vspace{-0.45cm}
\end{abstract}

\vspace{-0.6cm}
\section{Introduction}
\vspace{-0.2cm}

Multi-object tracking (MOT) is an indispensable component of applications such as autonomous driving \cite{Wang2018} and assistive robots \cite{Manglik2019}. Recent work approaches MOT in an online manner with a tracking-by-detection pipeline, where a detector \cite{Shi2019,Weng2019} is applied to all frames and features are extracted \emph{independently} from each object. Then, the pairwise feature similarity is computed between objects and used to solve MOT with a Hungarian algorithm \cite{WKuhn1955}. The key process of this pipeline is to learn discriminative features for objects with different identities.

\begin{figure}[t]
\vspace{-0.3cm}
\begin{center}
\includegraphics[trim=0cm 1.5cm 9.3cm 0cm, clip=true, width=0.7\linewidth]{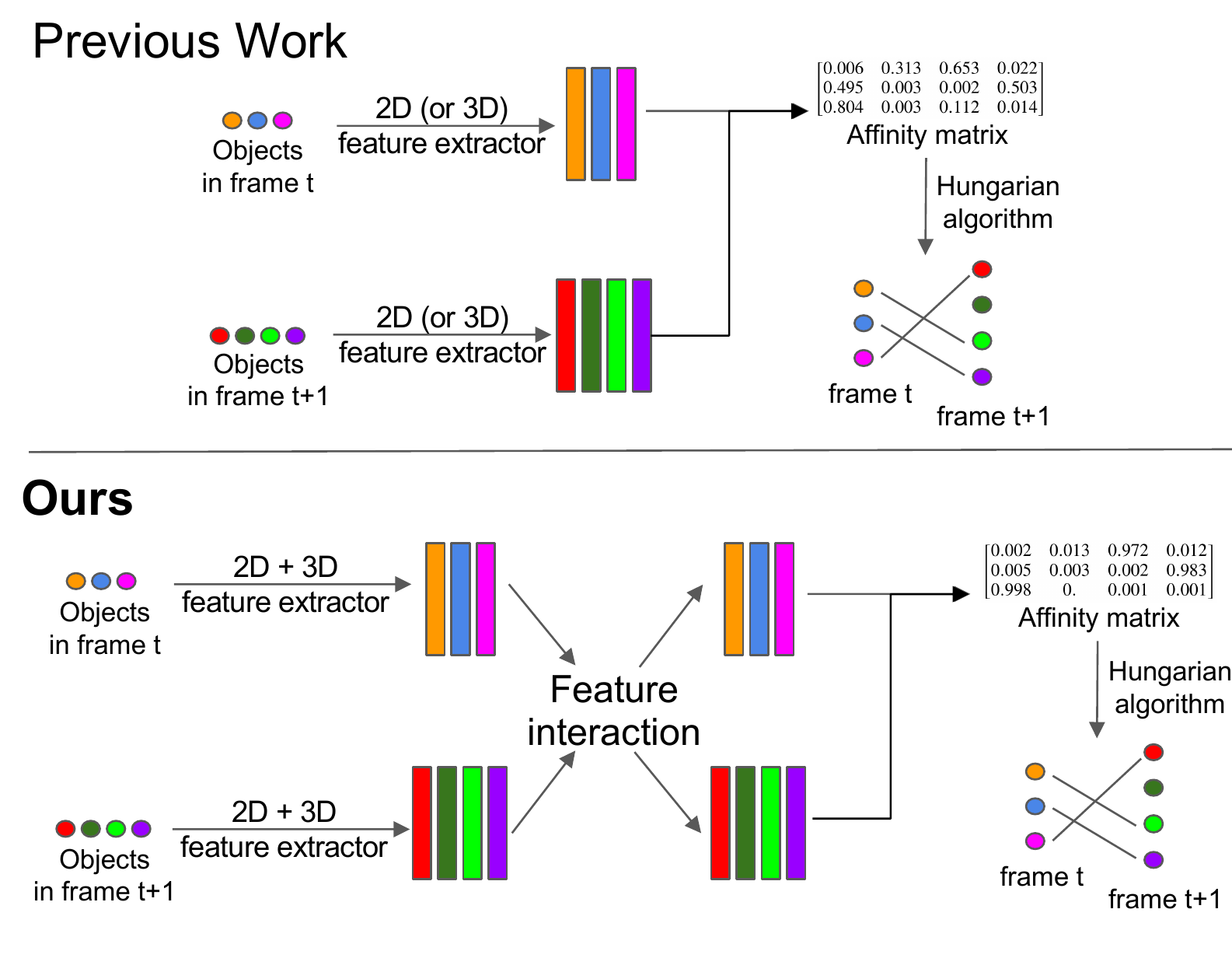}
\end{center}
\vspace{-1.3cm}
\caption{(\textbf{Top}): Prior work often employs a 2D or 3D feature extractor and obtain the feature independently from each object. (\textbf{Bottom}): Our work proposes a joint 2D and 3D feature extractor and a feature interaction mechanism to improve the discriminative feature learning for data association in MOT.}
\label{fig:teaser}
\vspace{-0.5cm}
\end{figure}
    
Our observation is that the feature extraction in prior work is independent for each object as shown in Figure \ref{fig:teaser} (Top) and there is no interaction. For example, an object's 2D appearance feature is computed only from its own image patch. We found that \textit{independent feature extraction leads to inferior discriminative feature learning}, and object dependency is the key to obtaining discriminative features. Intuitively, the features of the same object over two frames should be as similar as possible and the features between two different objects should be as different as possible to avoid confusion during data association. This can only be achieved if object features can be obtained as a dependent or context-aware process, i.e., modeling interactions between objects.

Based on the observation, we propose a novel \emph{feature interaction mechanism} for MOT as shown in Figure \ref{fig:teaser} (Bottom). We achieve this by introducing the Graph Neural Networks (GNNs). To the best of our knowledge, our work is the first applying the GNNs to MOT. Specifically, we construct a graph with each node being the object feature. Then, at every layer of the GNNs, each node can update its feature by aggregating features from other nodes. This node feature aggregation process is useful because each object feature is not isolated and can be adapted with respect to other objects. We observe that, after a few GNN layers, the computed affinity matrix becomes more discriminative.

In addition to the feature interaction, another critical aspect to discriminative feature learning in MOT is feature selection. Among different features, motion and appearance are proved to be the most useful features. Although prior works \cite{Baser2019} have explored using appearance and motion features, they only focus on 2D or 3D space as shown in Figure \ref{fig:teaser} (top). That means, prior work only uses the 2D feature when approaching 2D MOT or only uses the 3D feature for 3D MOT. However, this is not optimal as 2D and 3D information are complementary. For example, two objects can be very close in the image but actually at a distance in 3D space because of depth discrepancy. As a result, the 3D information is more discriminative in this case. On the other hand, 3D detection might not be very accurate for faraway objects which will result in noisy 3D motion. In this case, the 2D information might be more accurate for data association.

To this end, we also propose a novel feature extractor that jointly learns motion and appearance features from both 2D and 3D space as shown in Figure \ref{fig:teaser} (bottom). Specifically, the joint feature extractor has four branches with each branch being responsible for 2D appearance, 2D motion, 3D appearance and 3D motion feature, respectively. Features from all four branches are fused before feeding into the GNNs for feature interaction. 


\begin{figure*}[t]
\vspace{-0.2cm}
\begin{center}
\includegraphics[trim=0cm 0cm 0.2cm 0.1cm, clip=true, width=0.9\linewidth]{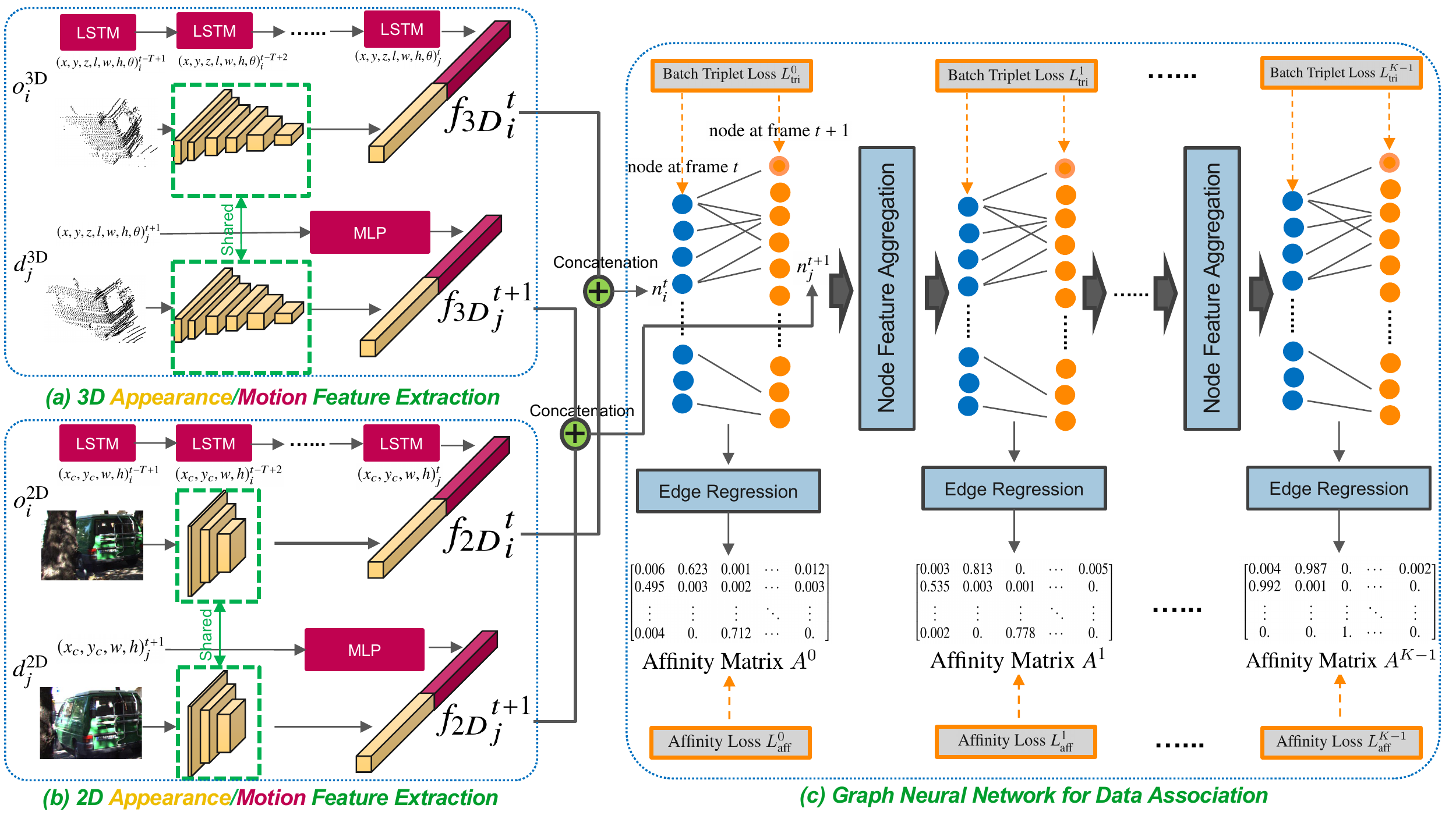}
\end{center}
\vspace{-0.8cm}
\caption{Proposed Network.} 
\label{fig:pipeline}
\vspace{-0.45cm}
\end{figure*}


\vspace{-0.5cm}
\section{Approach}
\vspace{-0.2cm}

The goal of online MOT is to associate existing tracked objects from previous frame with detected objects in the current frame. Given $M$ tracked objects $o_i \in O$ at frame $t$ where $i\in \{1, 2, \cdots, M\}$ and also $N$ detected objects $d_j \in D$ in frame $t$+$1$ where $j\in \{1, 2, \cdots, N\}$, we aim to learn discriminative features from $O$ and $D$ and find the matching based on the pairwise similarity. 

In Figure \ref{fig:pipeline}, our entire network consists of: (a) a 3D appearance and motion feature extractor; (b) a 2D appearance and motion feature extractor. Both 2D and 3D feature extractors are applied to all objects in $O$ and $D$ and then the extracted features are fused together, (c) a graph neural network that takes the fused object feature as inputs and constructs a graph with node being the object feature in the frame $t$ and $t$+$1$. Then, the GNNs iteratively aggregates the node feature from the neighborhood and computes the affinity matrix for matching using edge regression. To train the network, we employ the batch triplet loss on the node features and the affinity loss on the predicted affinity matrix.


\vspace{-0.5cm}
\section{Experiments}
\vspace{-0.2cm}

\noindent\textbf{Dataset.} To evaluate our 3D MOT method that requires both 2D and 3D data as inputs, we use the KITTI \cite{Geiger2012} dataset, which provides both the 2D (images and 2D boxes) and 3D data (LiDAR point cloud and 3D boxes). Same as most prior works, we primarily report results on the car subset for comparison. 

\vspace{1.5mm}\noindent\textbf{Evaluation Metrics.} We use standard CLEAR metrics \cite{Bernardin2008} (MOTA, MOTP, IDS, FRAG) and also the new sAMOTA, AMOTA and AMOTP metrics proposed in \cite{Weng2020_AB3DMOT} for MOT evaluation. Since we are evaluating 3D MOT methods, all above metrics need to be defined in 3D space using the criteria of the 3D IoU or 3D distance. However, the KITTI dataset only supports 2D MOT evaluation, i.e., metrics defined in 2D space for evaluating image-based MOT methods. Therefore, instead of using KITTI 2D MOT evaluation server, we use 3D MOT evaluation code released by \cite{Weng2020_AB3DMOT} for evaluation. Accordingly, evaluation must be done on the validation set as we do not have access to the ground truth of the test set for 3D MOT evaluation. As KITTI does not have an official train / validation split, we use the one proposed by \cite{Scheidegger2018}.

\vspace{1.5mm}\noindent\textbf{Baselines.} For 3D MOT, we compare with recent open-source 3D MOT systems such as FANTrack \cite{Baser2019}, mmMOT \cite{Zhang2019_robust} and AB3DMOT \cite{Weng2020_AB3DMOT}. To achieve a fair comparison, we use the same 3D detections obtained by PointRCNN \cite{Shi2019} for our proposed method and all baselines \cite{Baser2019,Zhang2019_robust,Weng2020_AB3DMOT} that require 3D detections as inputs. For baselines \cite{Baser2019,Zhang2019_robust} that also require 2D detections as inputs, we use the 2D projection of the 3D detections.

\begin{table*}[t]
\vspace{-0.15cm}
\caption{Quantitative comparison of 3D MOT performance on the KITTI dataset.}
\vspace{-0.3cm}
\centering
\resizebox{\textwidth}{!}{
\begin{tabular}{@{}llrrrrrrr@{}}
\toprule
Method \ \ \ \ \ \ \ \ \ \ \ \ & \textbf{sAMOTA}$\uparrow$ & \ \ AMOTA$\uparrow$ & \ \ AMOTP$\uparrow$ & \ \ MOTA$\uparrow$ & \ \ MOTP$\uparrow$ & \ \ IDS$\downarrow$ & \ \ FRAG$\downarrow$ \\
\midrule
mmMOT~\cite{Zhang2019_robust}   & 70.61 & 33.08 & 72.45 & 74.07 & 78.16 & 10 & 125\\ 
FANTrack~\cite{Baser2019}       & 82.97 & 40.03 & 75.01 & 74.30 & 75.24 & 35 & 202 \\
AB3DMOT\cite{Weng2020_AB3DMOT}  & 93.28 & 45.43 & 77.41 & \textbf{86.24} & 78.43 & \textbf{0} & 15  \\
\midrule
\textbf{Ours} & \textbf{93.92} & \textbf{45.83} & \textbf{78.10} & 86.03 & \textbf{79.03} & \textbf{0} &  \textbf{10} \\
\bottomrule
\end{tabular}}
\vspace{-0.5cm}
\label{tab:kitti_3d}
\end{table*}

\vspace{1.5mm}\noindent\textbf{Results.} We summarize the results in Table \ref{tab:kitti_3d}, where our method consistently outperforms other modern 3D MOT systems in most metrics. 


\vspace{-0.4cm}
\section{Conclusion}
\vspace{-0.2cm}

To improve discriminative feature learning, we proposed a new 3D MOT method with a novel joint feature extractor and a novel feature interaction mechanism achieved by GNNs. Through experiments on the KITTI dataset, we showed effectiveness of our method and established new S.O.T.A. 3D MOT performance.

%
%

\vspace{-0.4cm}
\bibliographystyle{splncs04}
\bibliography{main}
\end{document}